\documentclass[conference]{IEEEtran}
\IEEEoverridecommandlockouts
\usepackage{amsmath,amssymb,amsfonts}
\usepackage{algorithmic}
\usepackage{graphicx}
\usepackage{textcomp}
\usepackage{xcolor}
\usepackage{blindtext}
\usepackage{caption}
\usepackage{subcaption}
\usepackage[normalem]{ulem}
\def\BibTeX{{\rm B\kern-.05em{\sc i\kern-.025em b}\kern-.08em
    T\kern-.1667em\lower.7ex\hbox{E}\kern-.125emX}}

\usepackage[backend=bibtex,style=ieee]{biblatex}
\usepackage{fancyhdr}
\begin{document}

\title{Large Language Models in Wireless Application Design: In-Context Learning-enhanced Automatic Network Intrusion Detection\\
{
}
\vspace{-0.5em}}
\author{
    \IEEEauthorblockN{Han Zhang\IEEEauthorrefmark{1}, Akram Bin Sediq\IEEEauthorrefmark{2}, Ali Afana\IEEEauthorrefmark{2} and Melike Erol-Kantarci\IEEEauthorrefmark{1},\IEEEmembership{ Senior Member, IEEE}}
    
    \IEEEauthorblockA{\IEEEauthorrefmark{1} School of Electrical Engineering and Computer Science, University of Ottawa, Ottawa, Canada}
    
    \IEEEauthorblockA{\IEEEauthorrefmark{2} Ericsson Inc., Ottawa, Canada}
    
    \IEEEauthorblockA{\{hzhan363, melike.erolkantarci\}@uottawa.ca, }\IEEEauthorblockA{\{akram.bin.sediq, ali.afana\}@ericsson.com}\vspace{-3.5em}
}
\maketitle
\begin{abstract}
Large language models (LLMs), especially generative pre-trained transformers (GPTs), have recently demonstrated outstanding ability in information comprehension and problem-solving. This has motivated many studies in applying LLMs to wireless communication networks. In this paper, we propose a pre-trained LLM-empowered framework to perform fully automatic network intrusion detection. Three in-context learning methods are designed and compared to enhance the performance of LLMs. With experiments on a real network intrusion detection dataset, in-context learning proves to be highly beneficial in improving the task processing performance in a way that no further training or fine-tuning of LLMs is required. We show that for GPT-4, testing accuracy and F1-Score can be improved by 90\%. Moreover, pre-trained LLMs demonstrate big potential in performing wireless communication-related tasks. Specifically, the proposed framework can reach an accuracy and F1-Score of over 95\% on different types of attacks with GPT-4 using only 10 in-context learning examples.
\end{abstract}

\begin{IEEEkeywords}
Large language model, GPT, Generative AI, intrusion detection, wireless networks, B5G and 6G, in-context learning.
\end{IEEEkeywords}
\vspace{-15pt}
\section{Introduction} 
Artificial intelligence (AI) enables machines to perform automated perception and decision-making tasks that normally require human intelligence. In the past years, AI technologies have been applied to wireless communication applications to support a multitude of services. The widespread use of AI enables wireless networks to adapt to environmental changes in an automated fashion and serves as a basis for AI-native sixth-generation (6G) networks \cite{whitepaper}. 

Large language models (LLMs), especially generative pre-trained transformers (GPTs), have recently received extensive attention in various areas. Given the impressive capacity for information comprehension and strategic planning demonstrated by LLMs, there is a future vision of achieving a self-evolving wireless network \cite{bariah2023large} by integrating LLMs into Radio Access Network (RAN) controllers \cite{tarkoma2023ai}. The Generative AI journey in wireless networks begins by including LLM as a solution for wireless application design and developing LLM-driven network applications \cite{britto2023telecom}. 

Although the potential research directions of combining LLMs with wireless communications have been explored by several existing studies, only limited number of studies have proposed concrete realization methods or applications. The primary concern with this approach is that pre-trained LLMs may lack sufficient wireless communication-related knowledge to effectively execute complex tasks. Additionally, as models grow in scale, the process of fine-tuning LLMs becomes costly and technically challenging \cite{ding2023parameter}. On the other hand, in-context learning gives insights into tackling this challenge \cite{min2022rethinking}. In-context learning is a specific prompt-building method where task-related examples are provided as part of the prompt. In this way, it can empower the LLM domain-specific knowledge without updating the pre-trained model parameters.


There are several benefits of using pre-trained LLMs instead of traditional machine learning (ML) models. First, there is no need to train the ML models from scratch for each individual task in different scenarios. This provides the possibility of using one model to fit all the network functions since LLM parameters do not need to be updated when applied to different tasks. In addition, traditional ML-based methods usually require large amounts of training data to perform given tasks. In contrast, with the general intelligence obtained from pre-training, LLMs-based methods only need minimal example data. Using LLM also solves the over-fitting problems of traditional ML models. Moreover, the network robustness is improved since using pre-trained LLMs can prevent attacks during the model training phase. The explainability of model decisions can also be improved since LLMs can give semantic explanations about the decisions. Therefore, leveraging pre-trained LLMs for wireless network functions instead of traditional ML-based methods is a promising direction.

Inspired by the above-mentioned thoughts, in this work, we explore the potential of using pre-trained LLMs to automatically perform wireless communications-related tasks with in-context learning. Specifically, we propose a fully automated framework empowered by pre-trained LLMs and compare its performance to a traditional Convolutional Neural Network (CNN)-based network intrusion detection model. Three well-known LLMs are involved in the experiments, namely GPT-3.5, GPT-4, and LLAMA. Our focus is on evaluating the efficacy of LLMs in wireless communication tasks and the impact of prompt design. The main contributions are summarized as follows:

(1) We design a pipeline of automated LLM-empowered network applications. This pipeline involves LLM-based input information selection, prompt building, in-context learning, and output conversion. This realizes a seamless integration of LLMs within wireless communication systems.

(2) We propose a translation method between wireless network information and human-like text-based information to realize the communication between LLM and wireless networks.

(3) We propose three distinct in-context learning methods, namely illustrative in-context learning, heuristic in-context learning, and interactive in-context learning. Each is designed to significantly improve the performance of pre-trained LLMs in the specified task.

According to the experimental results, with the benefit of in-context learning, the pre-trained LLM is capable of performing the intrusion detection task. Specifically, for GPT-4, in-context learning can improve the detection accuracy and F1-Score by 90\%. With more than 10 in-context learning examples, GPT-4 can reach a detection accuracy and F1-Score higher than 95\%. It achieves comparable or superior performance to traditional ML models with only a small amount of task-specific data.

Although the proposed framework demonstrates good performance, LLM, being a very new technology, still have some potential pitfalls and risks while being applied to practical scenarios. Some frequently raised concerns include adversarial prompting, hallucination and stochastic output. These concerns can be mitigated by appropriate prompt design skills like output formatting and in-context learning. Still, more research on LLM needs to be conducted in the future to make it more reliable.

The rest of the paper is organized as follows. Section II introduces related works. Section III explains the implementation of the LLM-empowered network intrusion detection framework. Section IV describes the design of in-context learning methods. Section V gives experimental results, and Section VI concludes the paper.

\vspace{-4pt}
\section{Related works} 
There are a few studies that have applied pre-trained LLMs, especially GPT models, for domain-specific applications. For instance, \cite{wang2023voyager} proposed a framework using the GPT-4 model to explore the Minecraft world without human intervention. In \cite{wang2023chatcad}, medical domain knowledge and logical reasoning of LLMs are leveraged to enhance the output of computer-aided diagnosis networks. These studies have demonstrated the powerful capabilities of LLMs and have shown the potential for broader applications.

On the other hand, some studies have discussed how to use in-context learning to enhance the performance of LLMs. For instance, \cite{li2023large} discussed how to select effective in-context learning examples for LLMs to implement LLM-based code generation. In \cite{li2023overprompt}, in-context learning is applied to LLMs to handle multiple task inputs and reduce costs. Different from these works, our work  proposes three different ways to perform in-context learning and evaluate the effect of in-context learning on wireless communication-related tasks.

Some other works have discussed the combination of LLMs with wireless communications. In \cite{bariah2023large}, two aspects are discussed, which are how to leverage LLMs in wireless communications and how to empower wireless communication with LLMs. \cite{tarkoma2023ai} discussed several ways to integrate LLMs within the 6G system. These works are from the perspective of conducting an overview, however, our research can be seen as a concrete practice of possible research directions mentioned in previous works. In addition, \cite{jiang2023large} proposed a LLM-enhanced multi-agent system that can generate communication models with customized communication knowledge. \cite{shen2024large} proposed a LLM-based framework that can train and coordinate edge AI models to meet users' demands. Similarly, \cite{xu2024large} proposed
a split learning framework for LLM agents in 6G networks that enable collaboration between distributed devices. Our work uses LLMs in a different way. Instead of generating code with LLMs or designing generic frameworks, we directly generate wireless network-related decisions with logical reasoning.
It is worth noting that the proposed LLM framework is versatile and can be adapted for various network application designs beyond network intrusion detection. For instance, it can also be applied to cell configuration, incoming traffic analytics and dynamic resource management \cite{tarkoma2023ai}\cite{maatouk2023large}.
\vspace{-3pt}
\section{LLM-empowered network intrusion detection}
\begin{figure*}[!t]
\centerline{\includegraphics[width=5.4in]{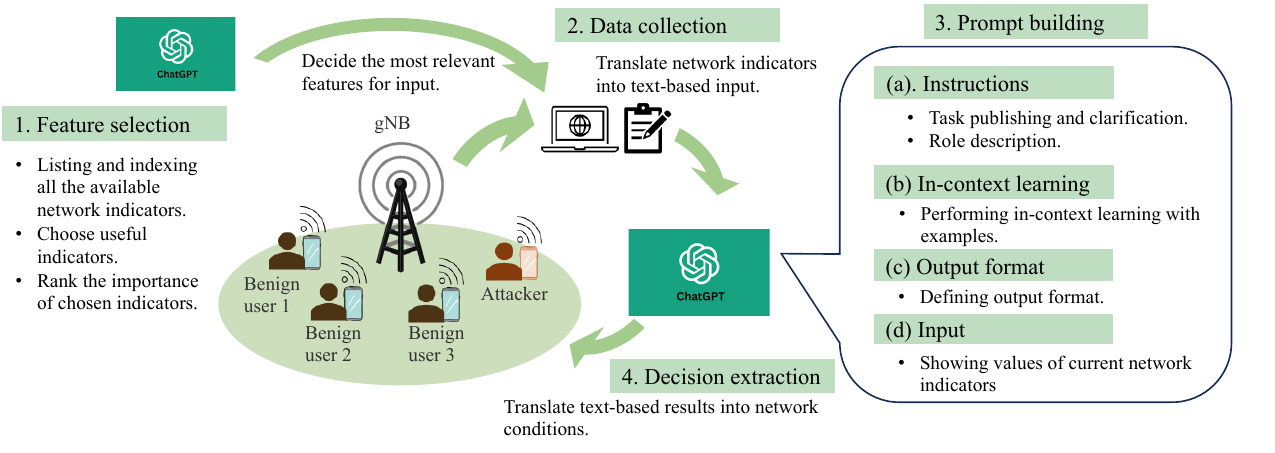}}
\captionsetup{font={small}}
\caption{System model of pre-trained LLM-empowered network intrusion detection. }
\label{fig1}
\vspace{-15pt}
\end{figure*}

The system model of the proposed pre-trained LLM-empowered network intrusion detection framework is shown in Fig \ref{fig1}. In this work, we consider a cloud-based wireless communication system where malicious attackers can perform distributed denial of service (DDoS) attacks and generate malicious traffic to the networks. At the network controller side, a cloud-based pre-trained LLM is deployed for network security monitoring and intrusion detection. 

As shown in Fig \ref{fig1}, four main steps are designed in the framework to enable fully automatic intrusion detection for zero-touch networks. In the first step, the most relevant network features for intrusion detection are selected by the LLM. In the second step, data is collected from the networks and processed for LLM input. The third step is to build the prompt for LLM, and the last step is to extract the desired decision from LLM output. These steps automate the interactions between the LLM and the 5G communication system, allowing the LLM to choose desired network information based on its knowledge base, capture the required values from the system, and feedback decisions. Detailed implementation of each step is explained below.

\vspace{-5pt}
\subsection{Feature selection}
To start LLM-based intrusion detection, feature selection is first performed to select the most important features from a large number of network features according to their relevance to the intrusion detection task. This step can be implemented with the knowledge base of pre-trained LLMs. 

First, all the accessible network features are indexed and given to the LLM. Next, the LLM is instructed to select the indexing numbers corresponding to the ten most relevant features of the network intrusion detection task. The LLM is then requested to rank the importance of all selected features on three levels: `very important', `kind of important', and `not very important'. Only 'very important' features and 'kind of important' features will be kept as desired features for the following detection tasks. The goal of this step is to concise the length of the LLM input token and to avoid the impact of irrelevant features on the detection results.
\vspace{-5pt}

\subsection{Data collection and processing}

After feature selection, the LLM is employed to monitor the network status for suspected intrusions. First, the values corresponding to the selected features are collected and converted to a text-based format. Next, a text-based template is created to give definitions and semantic explanations about the collected values. Finally, the text-based values are concatenated with pre-defined templates. Based on the above steps, selected input features can be translated into comprehensible human-like descriptions of current network conditions.
\vspace{-5pt}
\subsection{Prompt building}

In this step, prompts are designed to provide the LLM with formalized guidance for effective intrusion detection. The prompts are composed of four parts, instructions, in-context learning examples, output formatting, and input information. 

The instructions usually include the task publishing and clarification and role description. In particular, for network intrusion detection tasks, the prompts should first define what a network intrusion is, describe the role of the LLM as a 5G network safety monitor and clarify the task is to determine whether the traffic is from a malicious user based on given information. Next, examples of both benign cases and malicious cases are included in the prompts to provide the LLM with task-specific knowledge through in-context learning. The detailed implementation of in-context learning is given in the next section. 

Output formatting is also included as a part of prompt building to give clearer detection results and prevent misunderstandings. In particular, for network intrusion detection tasks, the output formatting can be implemented by using prompts like "If you think the traffic is malicious, answer yes. Otherwise if you think it is benign, answer no". The output formatting part also simplifies the translation from the LLM output to the detection results of network conditions in the following step. Input information refers to human-like descriptions of network conditions acquired in the last step.

\vspace{-5pt}
\subsection{Decision extraction}
\vspace{-5pt}
The last step is to understand the output of the LLM and extract the desired results. With output formatting in the last step, the decision extraction can be achieved with keyword searching. In detail, if "yes" or "Yes" appears in the LLM output while "no" or "No" does not appear, then the decision is that the traffic is malicious and vice versa. If neither of the above cases is valid, output formatting will continue to be performed until the desired clear output is obtained.
\vspace{-5pt}
\section{In-context learning-enhanced detection}
Although LLMs may have some basic intelligence and knowledge about wireless networks, it is still difficult to directly perform intrusion detection tasks through pre-trained LLMs without fine-tuning. In-context learning is an effective method that can improve the accuracy of LLMs  on specific tasks with only a small amount of labelled data. 

\begin{figure*}[!t]
\centerline{\includegraphics[width=5.0in]{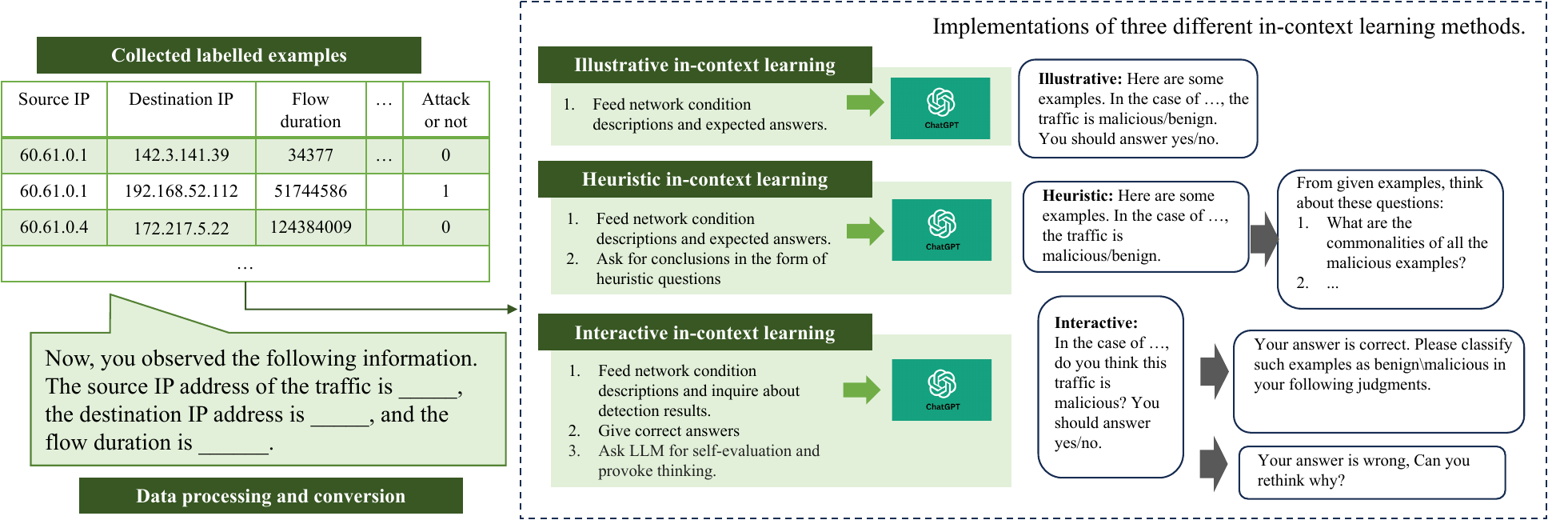}}
\captionsetup{font={small}}
\caption{Three different methods to implement in-context learning with labelled examples.}
\label{fig2}
\vspace{-15pt}
\end{figure*}

In this section, we illustrate how we design in-context learning schemes with labelled examples to improve the performance of pre-trained LLMs for domain-specific tasks. In first subsection, general principles of in-context learning in LLMs are introduced. In second subsection, detailed implementation of in-context learning in our framework is given.
\vspace{-5pt}
\subsection{In-context learning for LLM}
\vspace{-5pt}
In-context learning is an effective technique that enables pre-trained LLMs to address specific tasks without updating the parameters of LLM. This is achieved by integrating text-based examples into the prompts and by changing the distribution of the input text to obtain the desired output. The process can be formulated as:
\vspace{-10pt}
\begin{align}
    y_{t}^{*} = \underset{y_t}{argmax}\ P(y_{t}|(x_1, y_1), ...,(x_n, y_n), x_{t};\Theta_{origin})
\end{align}
where $y_{t}^{*}$ denotes the LLM output of the network intrusion detection task. $P(y_{t}|...)$ denotes the possibility of getting $y_{t}^{*}$ as the output. $x_{t}$ denotes the network condition description input. $(x_{n}, y_{n})$ denotes the $n^{th}$ in-context learning examples. $\Theta_{origin}$ denotes the pre-trained parameters of the LLM. Following this formula, the output of the LLM is influenced by examples included in the context information and the LLM is more likely to output decisions that are consistent with the given example \cite{li2023overprompt}.
\vspace{-5pt}
\subsection{In-context learning for network intrusion detection}
In this work, we design and compare three different in-context learning methods, the illustrative in-context learning, the heuristic in-context learning, and the interactive in-context learning. The implementations of three methods are presented in Fig. \ref{fig2}.

In the illustrative in-context learning method, examples with labels are converted into human (and LLM)-interpretable descriptions of cases and expected answers. Then, they are laid out and fed directly into the LLM as part of the prompt. The prompts usually start with identifying statements like "Here are some examples" and each example ends with conclusive instructions like "You should answer yes/no." 
\begin{figure*}[!t]
\centerline{\includegraphics[width=4.5in]{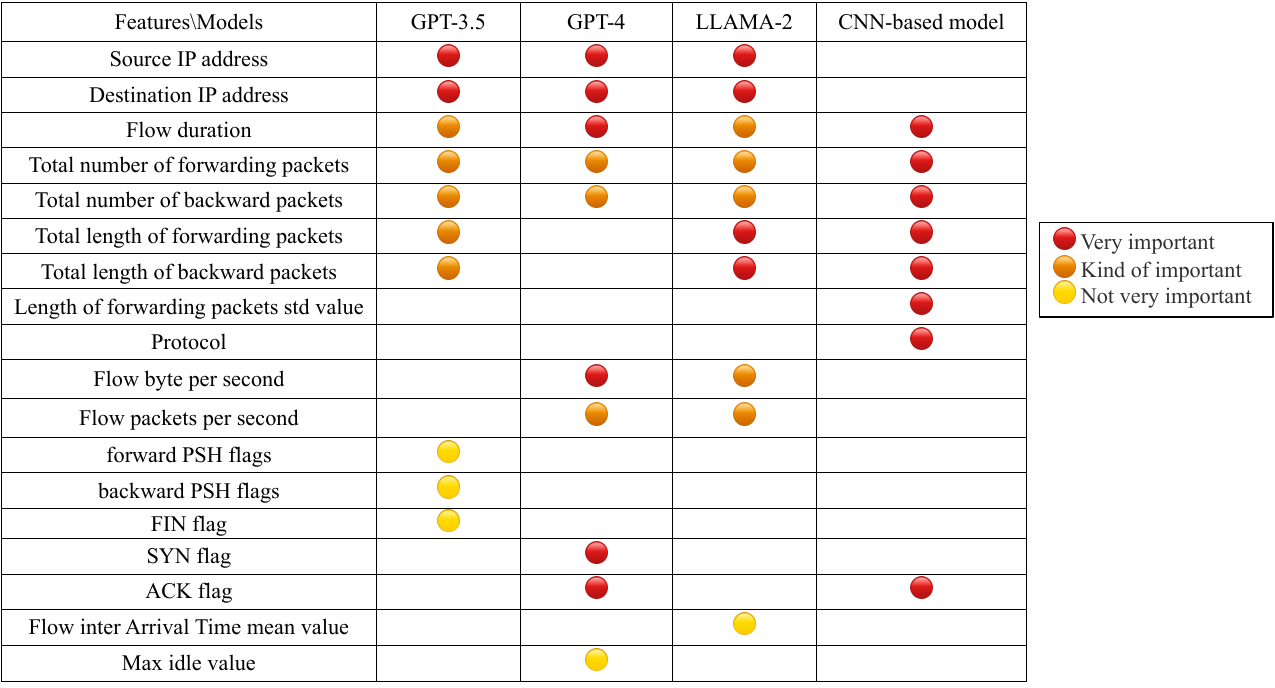}}
\captionsetup{font={small}}
\caption{The feature selection and importance ranking results from three different LLMs, GPT-3.5, GPT-4, and LLAMA. }

\label{fig3}
\vspace{-20pt}
\end{figure*}

The illustrative in-context learning method is the simplest among the three in-context learning methods. On this basis, the heuristic in-context learning method is designed by adding heuristic questions to the prompts. Based on the given network intrusion detection task scenario, some critical questions that may shape the outcome are extracted, like "What are the commonalities of all the malicious examples?" or "What is the rational range of flow duration?". These questions can give insights to the LLM on what should be concluded from the given examples. The LLM outputs of these questions will also be included in the prompts as context information for the following intrusion detection tasks. 

The last proposed in-context learning method is the interactive in-context learning. In this method, in-context examples are given and analyzed through a question-and-answer format. First, the examples are fed into the LLM without giving the expected answers. Then the LLM is asked to provide detection results and perform self-evaluation. If the result matches the label, the LLM is encouraged to continue to make judgments in this manner. Otherwise, the LLM is asked to perform self-corrections by providing explanations of expected results and reflecting on the reasons for the incorrect answer. Similar to the heuristic in-context learning method, the self-assessments and reflections will also be included in the prompts as a part of the context information.


\section{Experimental settings and Results}
\subsection{experimental settings}
In this work, we use a real network intrusion detection dataset proposed in \cite{farzaneh2023dtl} to generate testing examples and in-context learning examples. The dataset includes 9 types of DDoS attacks and 84 network traffic features. Three LLMs are used during the experiments, LLAMA-2-7b \cite{touvron2023llama}, GPT-3.5, and GPT-4 \cite{GPT}. CNN-based network intrusion detection is also implemented as a baseline and is used for the comparison between traditional ML models and LLMs. The performance is evaluated according to the accuracy and F1-Score of network intrusion detection. All the models are implemented using Python with the support of PyTorch, Openai-api, and huggingface.

\vspace{-5pt}
\subsection{experimental results}
\vspace{-2pt}
Fig \ref{fig3} shows the feature selection results of the network intrusion detection task from three LLMs, GPT-3.5, GPT-4, and LLAMA. The manually selected features for a CNN-based network intrusion detection model proposed in \cite{farzaneh2023dtl} are also added for comparison. During the experiments, each LLM is asked to choose ten features from the given 84 features, and the importance of these features are ranked according to three levels, 'very important', 'kind of important', and 'not very important'. Only 16 features are mentioned by three LLMs, and among these 16 features, 5 features mentioned by only one LLM are ranked as 'not very important'. This suggests that different LLMs exhibit a high degree of similarity in feature selection for the given task. It can also be observed that the selected features overlap significantly with the manually selected features. This validates the knowledge base of pre-trained LLM in the wireless communication domain and its ability to handle related tasks. In the following experiments, features ranked as 'not very important' are removed and only 'very important' features and 'kind of important' features are kept as the input for intrusion detection.

\begin{figure}[t]
\centering
 \begin{subfigure}[b]{0.3\textwidth}
     \centering
     \includegraphics[width=\textwidth]{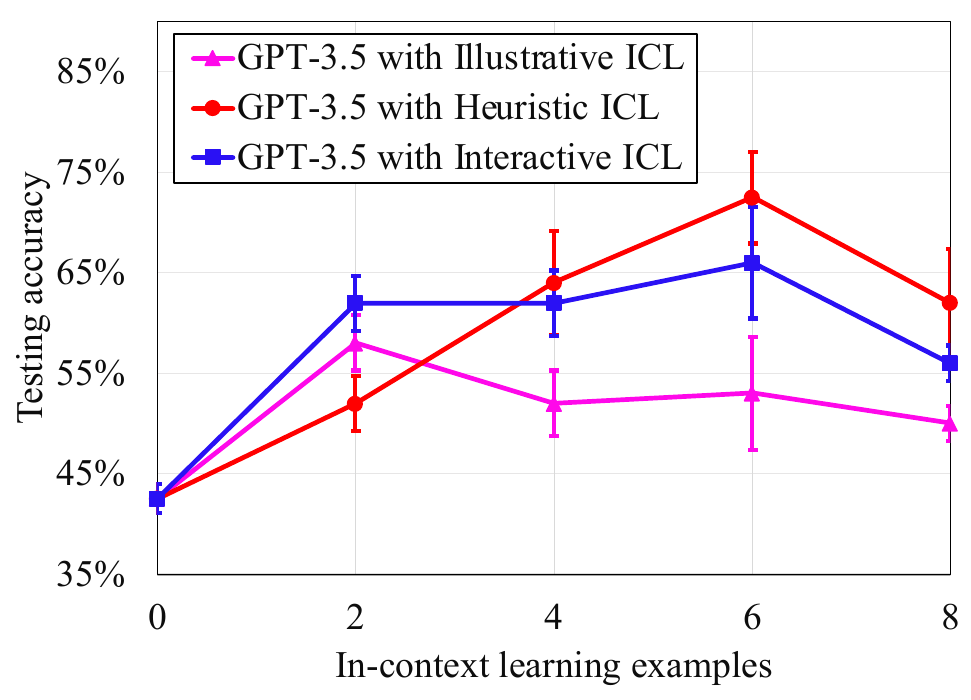}
     \vspace{-18pt}
     \captionsetup{font=scriptsize}
     \caption{Accuracy}
     \label{fig:y equals x}
 \end{subfigure}
  \hfill
 \begin{subfigure}[b]{0.3\textwidth}
     \centering
     \includegraphics[width=\textwidth]{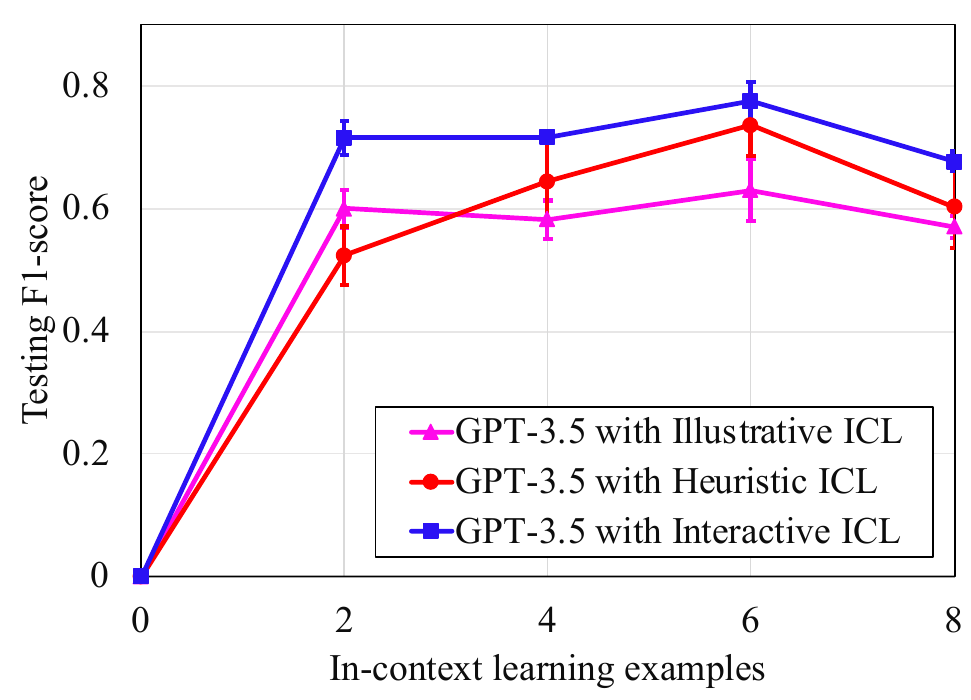}
     \vspace{-18pt}
     \captionsetup{font=scriptsize}
     \caption{F1-Score}
     \label{fig:three sin x}
 \end{subfigure}
 \vspace{-5pt}
 \captionsetup{font={small}}
\caption{Comparison of intrusion detection testing accuracy and F1-Score under three different in-context learning methods with GPT-3.5 while increasing the number of in-context learning examples.}
\label{fig4}
\vspace{-25pt}
\end{figure}

\begin{figure}[t]
\centering
 \begin{subfigure}[b]{0.32\textwidth}
     \centering
     \includegraphics[width=\textwidth]{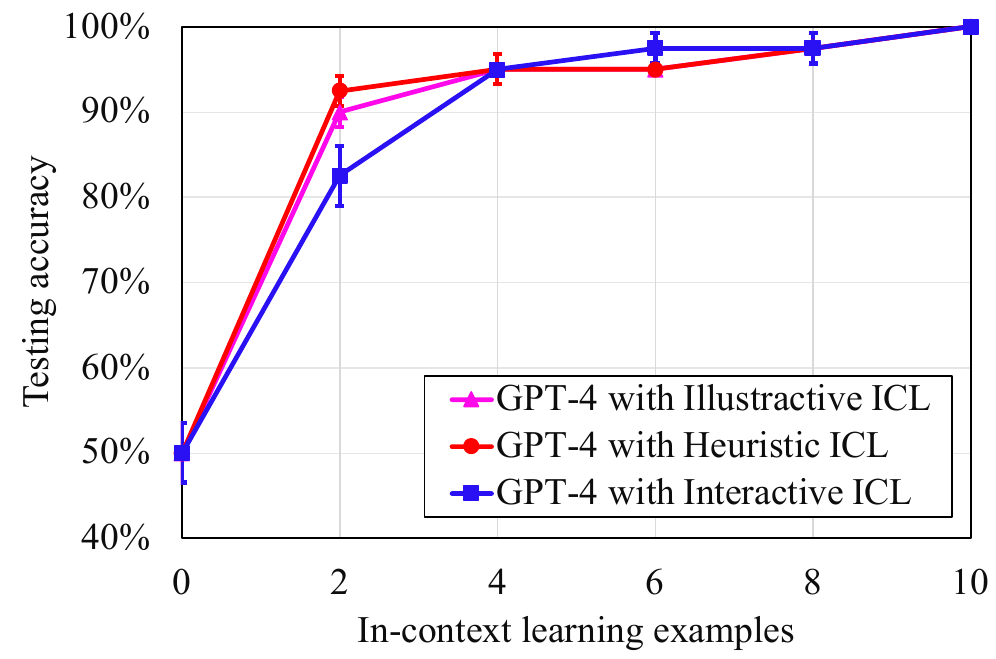}
     \vspace{-18pt}
     \captionsetup{font=scriptsize}
     \caption{Accuracy}
     \label{fig:y equals x}
 \end{subfigure}
  \hfill
 \begin{subfigure}[b]{0.32\textwidth}
     \centering
     \includegraphics[width=\textwidth]{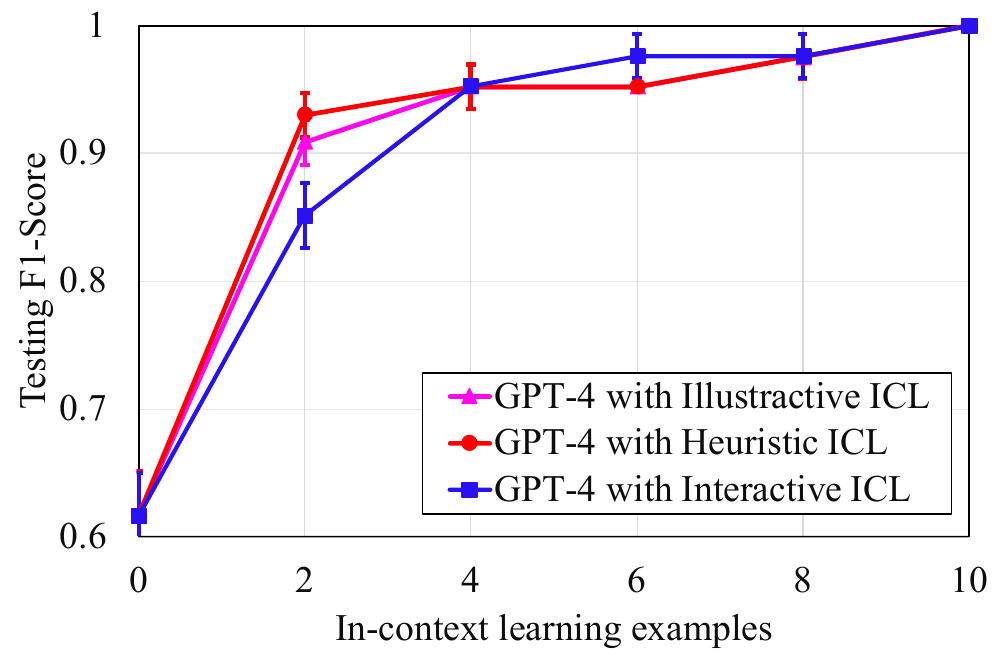}
     \vspace{-18pt}
     \captionsetup{font=scriptsize}
     \caption{F1-Score}
     \label{fig:three sin x}
 \end{subfigure}
 \vspace{-5pt}
 \captionsetup{font={small}}
\caption{Comparison of intrusion detection testing accuracy and F1-Score under three different in-context learning methods with GPT-4 while increasing the number of in-context learning examples.}
\label{fig4-1}
\vspace{-25pt}
\end{figure}

For the in-context learning experiments, we only test with GPT-3.5 and GPT-4. This is because LLAMA2 is an open-source model with a limited token size and is more suitable for fine-tuning rather than in-context learning. Fig. \ref{fig4} and Fig. \ref{fig4-1} show the comparison of three different in-context learning methods on a TCP ACK attack detection task. It can be observed that the testing accuracy and F1-Score grow with the number of examples. For GPT-3.5, the heuristic in-context learning method shows a poor performance when there are only 2 examples. This is because GPT-3.5 may summarize incorrect conclusions from only 2 examples. As a result, it may give wrong answers to heuristic questions and generate inaccurate detection results. When there are more in-context examples, the interactive in-context learning method and heuristic in-context learning method show better results than the illustrative in-context learning method. Furthermore, increasing the number of in-context learning examples does not necessarily lead to an improvement in performance. This is because of GPT-3.5 only has limited reasoning ability. When there are too many in-context examples, the model may fail to draw valid conclusions and show worse performance. It can also be observed that the accuracy and F1-Score show a slightly different trend. This is because with interactive in-context learning, GPT-3.5 will reflect on its mistakes and be more careful about possible intrusions. But this also makes it take the benign traffic as intrusions sometimes. So interactive in-context learning will lead to a relatively high F1-Score, but slightly lower accuracy.

Such observations do not appear in the experiments using GPT-4, which is equipped with a stronger reasoning ability. For GPT-4, the three different in-context learning methods lead to very close performance. The illustrative and the heuristic in-context learning methods perform a little bit better than the interactive in-context learning method when there are fewer examples. This is because GPT-4 has a greater contextual comprehension and reasoning ability \cite{liu2023evaluating}. As a result, it requires less guidance to complete the task.

Compared to illustrative in-context learning, heuristic in-context learning and interactive in-context learning will add extra information to the prompts and increase the token length. This will lead to higher inference costs. It is also worth noting that the performance of heuristic in-context learning depends partly on the design of the heuristic questions. Therefore, the design of heuristic in-context learning still requires the intervention of domain experts. This paper mainly focuses on the framework design and thus does not discuss the design of the heuristic questions in detail. In conclusion, illustrative in-context learning is a more cost-effective choice for LLMs with stronger capabilities. Heuristic in-context learning and interactive in-context learning will potentially offer performance boosts for LLMs with less capabilities.


\begin{figure}[t]
 \begin{subfigure}[b]{0.45\textwidth}
     \centering
     \includegraphics[width=\textwidth]{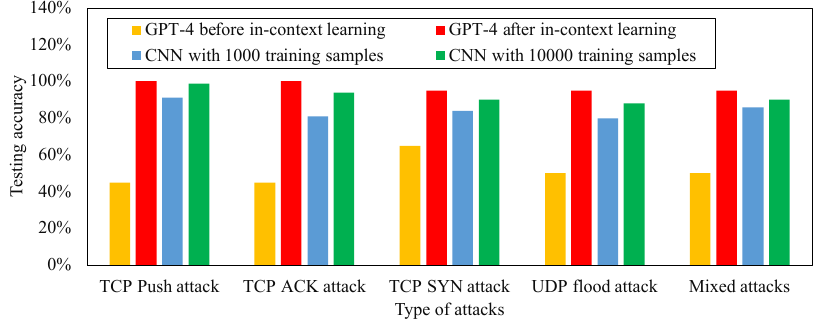}
     \vspace{-18pt}
     \captionsetup{font={small}}
     \captionsetup{font=scriptsize}
     \caption{Accuracy}
     \label{fig:y equals x}
 \end{subfigure}
  \hfill
 \begin{subfigure}[b]{0.45\textwidth}
     \centering
     \includegraphics[width=\textwidth]{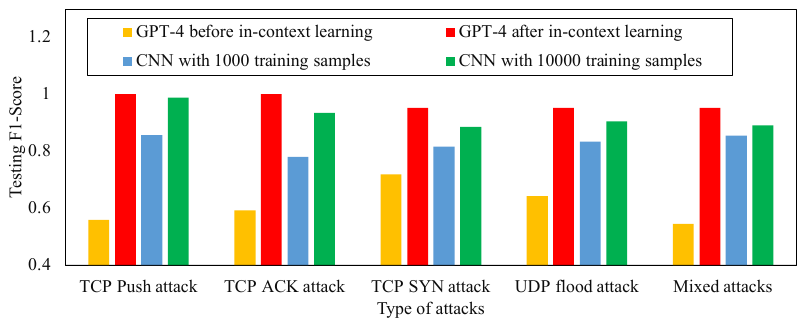}
     \vspace{-16pt}
     \captionsetup{font=scriptsize}
     \caption{F1-Score}
     \label{fig:three sin x}
 \end{subfigure}
 \captionsetup{font={small}}
 \vspace{-5pt}
\caption{Comparison of intrusion detection testing accuracy and F1-Score with GPT-4 and CNN-based method.}
\vspace{-20pt}
\label{fig6}
\end{figure}

Fig. \ref{fig6} compares the intrusion detection testing accuracy before and after in-context learning under different types of attacks. The experiments are performed by GPT-4 with 10 in-context learning examples. The performance of CNN-based intrusion detection with different numbers of training samples is also added for comparison. It can be observed that in-context learning can significantly improve the LLM performance on domain-specific tasks. The testing accuracy of LLM after in-context learning is about 90\% higher than before. On the other hand, the performance of the CNN model is very much related to the number of training samples. The accuracy and F1-Score will improve if the number of training samples increases. It can be concluded that the performance of traditional ML models relies much on massive amounts of data. In contrast, LLMs achieve comparable or superior performance to traditional ML models with only a small amount of task-specific data.
\vspace{-5pt}
\section{Conclusions and future directions}
\vspace{-5pt}
In this work, we investigated the potential of utilizing pre-trained LLMs for tasks within the realm of wireless communications. Specifically, we proposed a fully automatic network intrusion detection framework in which the LLM carries out the entire process, from feature selection to detection result output. We also proposed three ways to perform in-context learning with task-specific labelled data to enhance the performance. According to the experimental results, in-context learning can effectively improve the testing accuracy and F1-Score of LLMs by 90\%. After in-context learning, pre-trained GPT-4 model-based intrusion detection can reach an accuracy and F1-Score over 95\% on different types of intrusion attacks with only 10 in-context learning examples. 

This work is an initial attempt to apply LLMs to the field of wireless communication, and there are many research interests to delve into in the future. One thrust is on investigating how to choose more effective in-context learning examples and how to design heuristic questions for LLMs. Considering the potential LLMs have shown in this work, combining it with other wireless communication applications can be a promising direction.
\vspace{-10pt}
\section*{Acknowledgement}
 This work has been supported by NSERC Canada Research Chairs program, MITACS and Ericsson Canada.
\vspace{-5pt}

\normalem
\begin{refcontext}[sorting = none]
\small
\printbibliography
\end{refcontext}

\end{document}